\documentclass[conference]{IEEEtran}
\IEEEoverridecommandlockouts
\usepackage[compress]{cite}

\usepackage{amsmath,amssymb,amsfonts}
\usepackage{algorithmic}
\usepackage{graphicx}
\usepackage{textcomp}
\usepackage{xcolor}
\usepackage{makecell}
\usepackage{booktabs}
\usepackage{multirow}
\usepackage{tabularx}
\usepackage{siunitx}  % For aligning numbers on the decimal point
\usepackage{hyperref}
\usepackage[colorinlistoftodos,prependcaption,textsize=tiny]{todonotes}
\usepackage{threeparttable} % for table notes
\def\BibTeX{{\rm B\kern-.05em{\sc i\kern-.025em b}\kern-.08em
    T\kern-.1667em\lower.7ex\hbox{E}\kern-.125emX}}
\begin{document}

\title{JobHop: A Large-Scale Dataset of Career Trajectories\\
\thanks{This research was funded by the BOF of Ghent University (BOF20/IBF/117), 
the Flemish Government (AI Research Program), the FWO (G0F9816N, 3G042220, G073924N), 
and the EU (ERC, VIGILIA, 101142229). 
Views and opinions expressed are however those of the author(s) only and do not necessarily reflect those of the EU or the ERC Executive Agency. 
Neither the EU nor the granting authority can be held responsible for them. 
Part of the experiments were conducted on pseudonymised HR data generously provided by VDAB. 
The authors also thank Guillaume Bied for his valuable input and support, and Marybeth Defrance for her help in annotating the datasets.
}
}

\author{\IEEEauthorblockN{Iman Johary}
\IEEEauthorblockA{\textit{IDLab} \\
\textit{Ghent University}\\
Ghent, Belgium \\
iman.johary@ugent.be}
\and
\IEEEauthorblockN{Raphaël Romero}
\IEEEauthorblockA{\textit{IDLab} \\
\textit{Ghent University}\\
Ghent, Belgium \\
raphael.romero@ugent.be}
\and
\IEEEauthorblockN{Alexandru C. Mara}
\IEEEauthorblockA{\textit{IDLab} \\
\textit{Ghent University}\\
Ghent, Belgium \\
alexandru.mara@ugent.be}
\and
\IEEEauthorblockN{Tijl De Bie}
\IEEEauthorblockA{\textit{IDLab} \\
\textit{Ghent University}\\
Ghent, Belgium \\
tijl.debie@ugent.be}
}

\maketitle

\begin{abstract}
Understanding labor market dynamics is essential for policymakers, employers, and job seekers.
However, comprehensive datasets that capture real-world career trajectories are scarce.
In this paper, we introduce JobHop, a large-scale public dataset derived from anonymized resumes provided by VDAB, the public employment service in Flanders, Belgium. 
Utilizing Large Language Models (LLMs), we process unstructured resume data to extract structured career information, which is then normalized to standardized ESCO occupation codes using a multi-label classification model. 
This results in a rich dataset of over 1.67 million work experiences, extracted from and grouped into more than 361,000 user resumes and mapped to standardized ESCO occupation codes, offering valuable insights into real-world occupational transitions.

This dataset enables diverse applications, such as analyzing labor market mobility, job stability, and the effects of career breaks on occupational transitions.
It also supports career path prediction and other data-driven decision-making processes.
To illustrate its potential, we explore key dataset characteristics, including job distributions, career breaks, and job transitions, demonstrating its value for advancing labor market research.
\end{abstract}


%===================================================
%===================================================
\section{Introduction}\label{sec_intro}
%===================================================
%===================================================

Understanding the complex dynamics of the labor market—shaped by economic conditions, technological advancements,
and social trends—is essential for policymakers, employers, and job seekers \cite{RAHHAL2024124101}.
Insights into labor market trends guide policy development to address skill shortages and unemployment \cite{vankevich_ensuring_2020},
enhance hiring efficiency and employee retention for employers \cite{math11224677},
and improve career guidance and employment opportunities for job seekers \cite{alibasic_evaluation_2022}.

The increasing availability of labor market data,
combined with advancements in machine learning and natural language processing,
presents new opportunities for data-driven analyses and decision-making \cite{horton2015labor}.
However, publicly accessible large datasets capturing real-world occupational trajectories remain scarce,
limiting the scope of labor market research and applications such as career path prediction, workforce planning, and policy analysis.

This paper introduces JobHop,
a large-scale dataset derived from anonymized resumes provided by VDAB,
the public employment service of Flanders, Belgium.
The dataset includes ESCO occupation codes with corresponding dates for over 1.67 million work experiences from more than 361,000 resumes,
offering insights into occupational transitions.
This dataset was created by using Large Language Models (LLMs) to extract structured career information from unstructured textual data 
                     and by mapping the occupations to ESCO codes via a multi-label classification model.

The standardized ESCO taxonomy enables detailed analysis of career trajectories, mobility patterns, job stability,
and the influence of education and career breaks on occupational transitions.
It also facilitates international comparisons and integration with other labor market data sources.
Beyond labor market analysis,
the dataset supports applications such as career path prediction, skill gap analysis, workforce planning,
and data-driven career guidance tools.

To demonstrate the potential of the dataset,
we explore several aspects of the Flemish labour market such as the impact of career breaks on job transitions,
mobility within the job market, and job stability across different sectors.
We also study the influence of university degrees on career trajectories,
offering insights into tertiary attainments and labor market integration.

As a key contribution to labor market research and career path prediction,
we release this aggregated dataset publicly\footnote{The dataset is available at: \href{https://huggingface.co/datasets/aida-ugent/JobHop}{huggingface.co/datasets/aida-ugent/JobHop}}.
The dataset is designed to be regularly updated to reflect advancements in LLM technology and to incorporate new data,
ensuring its continued relevance and value to researchers, policymakers, and practitioners.

The main contributions of this paper are as follows:

\begin{itemize}
	\item We introduce JobHop, a comprehensive dataset of real-world ESCO occupation transitions,
	 enabling a wide range of labor market analyses and career path prediction applications.
	\item We present an LLM-based pipeline for extracting structured career information from unstructured resume data and mapping it to standardized ESCO occupation codes.
	\item We demonstrate the potential of the dataset through analysis of the Flemish labor market, 
	showcasing its versatility and value for data-driven decision-making.
\end{itemize}

The remainder of this paper is structured as follows.
Sec.~\ref{sec_related} reviews related work.
Sec.~\ref{sec_processing} details the data curation process from structural information extraction to data normalization.
Sec.~\ref{sec_analysis} provides an analysis of job market's trends and patterns, showcasing the dataset's potential.
Finally, Sec.~\ref{sec_conclusion} concludes the paper and discusses avenues for future work.

%===================================================
%===================================================
\section{Related Work}\label{sec_related}
%===================================================
%===================================================

This section first provides an overview of currently available datasets describing career trajectories.
We then provide background on Named Entity Recognition and Occupation Classification methods,
setting the stage for the explanation of choices made for dataset curation in Sec.~\ref{sec_processing}.

%===================================================
\subsection{Labor Market Datasets\label{ssec_related_ner}}
%===================================================

Gathering and publicly sharing large-scale data on job trajectories is costly and raises privacy concerns.
Long-running panel surveys,
such as the eight panel surveys combined in the Cross-National Equivalent File \cite{CNEF},
as well as the National Longitudinal Surveys (NLSY) \cite{moore2000national},
the China Family Panel Studies (CFPS) \cite{xie2014introduction},
the LISS \cite{scherpenzeel2018true},
the SIPP \cite{us2004survey},
and the LLMDB \cite{LLMDB},
provide crucial information but typically represent very small sample sizes (relative to the millions of profiles available online) 
and often require complex procedures for research access or registration.
National or international Labor Force Surveys \cite{donovan2023labor} and the U.S.
Current Population Survey\cite{cps2023} usually adopt rotating panel structures 
(where different groups of respondents are surveyed for a few periods before being replaced by new groups),
which limit their ability to capture long-term dynamics at the individual level.

Accordingly, our work addresses a gap identified by Senger et al. \cite{karrierewege} in datasets used for predicting career paths in machine learning.
Aside from the use of a rotating survey by Chang et al. \cite{chang2019bayesian}, in which individuals are observed for only two years,
most studies have focused on learning from proprietary datasets created from online platforms such as LinkedIn \cite{li2017nemo, cerilla2023career},
CareerBuilder \cite{dave2018combined}, or Zippia \cite{vafa2022career}.
These studies often provide evaluation results on small yet higher-quality panels \cite{vafa2022career, du2024labor, chang2019bayesian}.
The evaluations frequently suggest the complementarity between large,
yet potentially less curated,
proprietary datasets and high-quality survey data for modeling and analysis.

Smaller public resume datasets are also available.
Decorte et al. \cite{decorte2023career} provide a dataset containing 2,164 anonymized,
publicly available resumes sourced from the Kaggle platform.
The OpenResume dataset\cite{openresume} combines 301 real resumes with 3,000 synthetically generated resumes, totaling 3,301 entries.
In addition to \cite{karrierewege},
which presents a large-scale ESCO occupation transition dataset sourced from the German Employment Agency,
this work aims to offer a large-scale and public dataset of job transitions with dates.
This dataset can be used for studying career trajectories,
labor market dynamics,
and as a benchmark for career path prediction.

\begin{table}[htbp]
\caption{Comparison of datasets describing career trajectories.}
\resizebox{\columnwidth}{!}{
\begin{tabular}{l c c c}
\toprule
\textbf{Dataset} & \textbf{Sample Size} & \textbf{Data Type} & \textbf{Access} \\
\midrule
CNEF \cite{CNEF} & 300k+ & Survey & Restricted \\
NLSY \cite{moore2000national} & 12k+ & Survey & Restricted \\
CFPS \cite{xie2014introduction} & 30k+ & Survey & Public (reg.) \\
LISS \cite{scherpenzeel2018true} & $\sim$7k & Survey & Restricted \\
SIPP \cite{us2004survey} & 50k+ & Survey & Restricted \\
LLMDB \cite{LLMDB} & $\sim$600k & Admin. records & Restricted \\
LinkedIn \cite{li2017nemo, cerilla2023career} & 100M+ & Online resume & Proprietary \\
CareerBuilder \cite{dave2018combined} & 10M+ & Online resume & Proprietary \\
Zippia \cite{vafa2022career} & 10M+ & Online resume & Proprietary \\
DECORTE \cite{decorte2023career} & 2,164 & Resume & Public \\
OpenResume \cite{openresume} & 3,301 & Resume & Public \\
Karrierewege \cite{karrierewege} & 500k+ & ESCO code & Public \\
\textbf{JobHop} & 361k+ & ESCO code + dates & Public \\
\bottomrule
\end{tabular}
}
\begin{tablenotes}
\footnotesize
\item \textit{(reg.)}:  short for registration required.
\item \textit{Restricted}: available only upon formal request or under license (e.g., CNEF, LISS).
\end{tablenotes}
\label{tab:dataset_comparison}
\end{table}

%===================================================
\subsection{Dataset Curation \label{ssec_dataset_curation}}
%===================================================

%===================================================
\subsubsection{Named Entity Recognition \label{ssec_related_ner}}
%===================================================

Named Entity Recognition (NER) is a sub-task of information extraction that aims to identify specific entities in text and classify them into predefined categories.
NER has been utilized for extracting information from resumes \cite{italian_resume_extraction, resume_extraction,resume_parsing}.
These works involve segmenting resumes into smaller,
semantically similar subsections and training different NER models for each subsection.
These approaches, however, generally rely on supervised models which require a large amount of labeled data.

Large language models (LLMs) \cite{llama2, mistral}, capable of diverse NLP tasks,
have shown promising results in named entity recognition (NER) \cite{llm_extraction, llm_medical_extraction}.
Although their NER performance may lag behind that of supervised models, LLMs require little or no labeled data.
They can identify entities without prior examples through zero-shot prompting \cite{NEURIPS2022_8bb0d291} or with a few examples, known as few-shot prompting.
Given the dynamic nature of job market data and the variability in resume quality, developing comprehensive NER models is challenging.
LLMs offer a flexible, adaptable solution to address this complexity.

%===================================================
\subsubsection{Occupation Classification \label{ssec_related_ESCO}}
%===================================================

The classification of job postings into standard occupation taxonomies such as ESCO is a fundamental task in labour market analysis.
Traditionally, this classification was performed through supervised algorithms leveraging exclusively job title information
 \cite{job_classification,job_classification2,job_classification4}.
In contrast, Varelas et al. \cite{job_classification5} 
propose an ensemble of five supervised machine learning models that utilize both job titles and job descriptions.
The resulting framework was used to classify Greek job postings into ESCO categories.
These supervised models, 
however, require large amounts of labeled data for training, 
which are costly to produce and, thus, rarely available.

More recently, Li et al. \cite{llm4jobs} introduced a new model for unsupervised classification of job postings based on LLMs.
This model combines job titles and job descriptions into vector representations 
which are compared to the representations of ESCO occupations to retrieve the appropriate codes.
The model shows promising results and, unlike supervised approaches, does not require large amounts of labeled data.

%===================================================
%===================================================
\section{Dataset Curation}\label{sec_processing}
%===================================================
%===================================================

In this section, 
we describe the process of transforming unstructured resume data into a structured format that captures work experiences and educational qualifications. 
We introduce the resulting dataset, JobHop, which contains over 1.67 million work experiences with dates. 
The dataset creation process involves two steps: the first step converts unstructured resumes into structured data, 
and the second step normalizes the structured data by mapping job titles to ESCO codes. 
In this section, we discuss the three phases of the dataset: raw resumes, structured dataset, and normalized dataset.
In Sec.~\ref{ssec_data_overview}, we provide an overview of the resumes used to create JobHop. 
Then, we discuss the process of converting raw resumes into a structured dataset (Sec.~\ref{ssec_info_extraction}). 
Lastly, we describe the normalization process, 
which maps job titles to ESCO codes, resulting in the final normalized "JobHop" dataset (Sec.~\ref{ssec_mapping}).

\subsection{Data Overview}\label{ssec_data_overview}
%===================================================

To investigate variations in resume content across different sectors, backgrounds, and individuals, 
we utilized a proprietary dataset comprising 442,555 resumes from 385,052 unique individuals, 
provided by the Flemish Public Employment Service (VDAB). 
Prior to sharing the data, 
VDAB converted all resumes into plain text and anonymized them by removing personally identifiable information such as names and addresses. 
Additionally, terms with low Term Frequency–Inverse Document Frequency (TF-IDF) were excluded to minimize the risk of indirect identification.
% Prior to sharing the data, data we converted into plain text and,
% VDAB anonymized all resumes and removing personally identifiable information such as names and addresses. 
% Additionally, terms with low Term Frequency–Inverse Document Frequency (TF-IDF) were excluded to reduce the risk of indirect identification. 
This approach, while omitting some specific company names, job titles, or educational institutions, 
retained sufficient content for meaningful analysis. 

The multilingual corpus is predominantly in Dutch (91.4\%), with smaller portions in English (6.1\%) and French (2.4\%), 
covering professional experience, educational background, skills, and other relevant attributes. 
Following initial inspection, 
we removed resumes that had become uninformative after anonymization or contained format conversion errors rendering the text unintelligible.
After this and other filtering steps described in Sec.~\ref{ssec_info_extraction},
the final dataset comprised 361{,}207 resumes suitable for analysis

\emph{Ethical use of data:} The resumes originate from candidates seeking employment through VDAB. 
As part of a research collaboration with VDAB, 
we are authorized to process this data for research purposes in a manner that strictly adheres to ethical and legal requirements.
The resumes were already anonymized to protect individual privacy.
All data processing and analysis were conducted internally, with no data shared with third parties.

The released dataset, JobHop, is presented in highly aggregated form, containing no personally identifiable information.
Additionally, the released dataset underwent further anonymization measures, 
including the removal of location data and the use of fixed granularity, where all dates were rounded to the nearest quarter to ensure privacy.

\subsection{Extracting Experiences and Qualifications}\label{ssec_info_extraction}
%===================================================

This subsection describes the first step of our pipeline: 
extracting structured work experiences and academic qualifications from unstructured resumes, which vary widely in format and detail.
The information is extracted in a machine-readable JSON format for further analysis.

From each resume, we extract information related to work experiences, including job title, job description, company name, start and end dates, and location. 
Additionally, we extract academic qualifications in the form of degree, institution name, and study period.
The structured data includes work experiences, 
covering both employment and internships, 
and qualifications, 
encompassing education and certifications.

\begin{figure}[htbp]
	\centering
	\resizebox{\columnwidth}{!}{
	   \includegraphics[width=\linewidth]{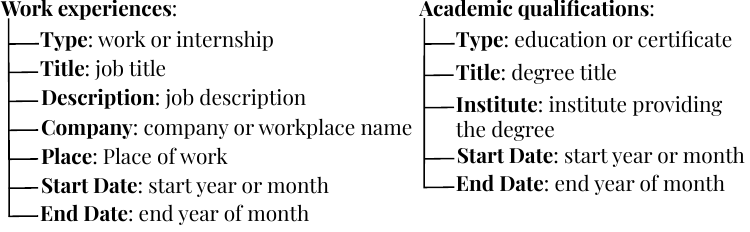}
	}
	\caption{Structured output format expected to be returned by the LLM.}
	\label{fig:json_structure}
\end{figure}

Large Language Models (LLMs) were used to extract relevant information efficiently. 
Their ability to identify complex relationships within lengthy texts \cite{kuratov2024search} and to handle noise from pseudonymization makes them well-suited for this task. 
By providing predefined instructions to output JSON, 
unstructured resumes are transformed into structured data ready for data analysis pipelines.
Figure~\ref{fig:json_structure} illustrates the JSON structure produced by the LLM.

To ensure consistent and accurate extraction, 
various prompt engineering strategies were tested. 
Due to the length of resumes, 
one-shot prompting was used to remain within context length constraints while minimizing hallucinations and improving consistency. 
Refining the example prompt to address common extraction errors further improved reliability. 
Experiments with different methods for structural extraction revealed that one-shot prompting combined with JSON output formatting provided optimal results.

Table~\ref{tab:prompt_structure} presents the prompt structure used. 
The system message is divided into four parts: 
defining the role, 
explaining the input and output, 
specifying the output format, 
and providing instructions. 
The first user message includes a one-shot input example, 
and the assistant’s message shows the expected JSON output. 
The second user message contains the unstructured resume, to which the LLM responds with structured data in the expected format.

\begin{table}[htbp]
	\caption{Overview of the proposed prompt structure.}
	\resizebox{\columnwidth}{!}{
	\begin{tabular}{p{1cm} p{2.5cm} p{4cm}}
		\toprule
		\textbf{Role} & \textbf{Prompt Section} & \textbf{Goal} \\
		\midrule
		\multirow{4}{*}{SYSTEM} 
			& Defining Role & Instruct LLM to act as HR assistant \\
			& Input/Output & Clarify expected input and output \\
			& Output Format & Specify output format \\
			& Instructions & Additional instructions \\
		\midrule
		USER & Example Resume & One-shot input example \\
		\midrule
		MODEL & Extracted Info & Expected output for example input \\
		\midrule
		USER & Unstructured Resume & Resume to be structured \\
		\bottomrule
	\end{tabular}
	}
	\label{tab:prompt_structure}
\end{table}

After extracting information from resumes, 
we consolidated consecutive work experience entries with identical job titles, company names, and job descriptions. 
Additionally, we filtered out resumes that listed more than 20 work experiences and had less than 80\% unique dates among their work experiences, 
as these often contained extraction errors or irrelevant information.
By applying these filtering criteria, we obtained the final dataset of 361,207 resumes.

\emph{Validation:}\label{sec_eval_structural_extraction} To evaluate the extraction of work experiences and qualifications, 
we hand-annotated a dataset consisting of 200 resumes.
The annotation process, carried out by the authors with some external assistance, took approximately 60 hours.
Performance was measured by calculating similarity scores between the extracted fields and the ground truth. 
Each field received a score between 0 and 1, with missing fields assigned a score of 0. 
We then computed the average score for each resume. 
For text fields, we applied fuzzy string matching using the FuzzyWuzzy library, 
while date fields were evaluated using binary scoring, assigning 1 for exact matches and 0 otherwise.

Table~\ref{tab:model_comparison} summarizes the results. 
JSON accuracy indicates the percentage of outputs that can be successfully converted to JSON, 
while title accuracy measures the average accuracy across extracted job and degree titles. 
Overall accuracy refers to the average similarity score per resume. 
Among the evaluated models, Gemma2-9b achieved the highest JSON and overall accuracy.
It is worth noting that these accuracies reflect the agreement between the models and human annotators; perfect agreement is not expected, as even human inter-annotator agreement is typically below 100\%.

We also examined the impact of zero-shot versus one-shot prompting using Vicuna-13b. 
One-shot prompting significantly improved performance, 
increasing overall accuracy from 57.9\% to 71.5\% and JSON accuracy from 89\% to 91\%. 
Consequently, all subsequent experiments used one-shot prompting.

\begin{table}[htbp]
\caption{Comparison of different LLMs based on JSON accuracy, title accuracy, and overall accuracy.}
\begin{center}
\resizebox{\columnwidth}{!}{
\begin{tabular}{|c|c|c|c|}
\hline
\textbf{Model} & \textbf{\textit{JSON Acc}} & \textbf{\textit{Title Acc}} & \textbf{\textit{Overall Acc}} \\
\hline
Vicuna-13b & 91 & 72.7 & 71.5 \\
\hline
Llama3-8b it \cite{llama3} & 99 & 75.9 & 75.5 \\
\hline
Llama3.1-8b it \cite{llama3} & 98 & 78.7 & 78.9 \\
\hline
Gemma2-9b it \cite{gemma2} & 99.5 & 81.6 & 82.1 \\
\hline
\end{tabular}
}
\label{tab:model_comparison}
\end{center}
\end{table}

Due to the dataset size and the computational demands of these models, 
converting the dataset into structured data required approximately 500 GPU days\footnote{
Data extraction and validation were performed on multiple machines totaling 14 Nvidia A40 GPUs, 1 TB of RAM, and 48 CPU cores.
} on our current infrastructure. 
Educational and employment details were extracted to analyze the impact of tertiary education on occupation distribution.

%===================================================
\subsection{Data Normalization\label{ssec_mapping}}
%===================================================

Individuals often describe their job titles in varying ways, 
making it challenging to align them under a common language that enables straightforward analysis.
To address this, we employ the European Skills, Competences, Qualifications, and Occupations (ESCO) taxonomy, 
which is a comprehensive taxonomy used to categorize occupations based on skill level, specialization, and requirements.
ESCO facilitates standardized comparisons of labor markets across regions and countries, 
aiding governments, organizations, and researchers in analyzing labor market dynamics.

ESCO is structured hierarchically, 
with ten broad groups at the first level, 
referred to as \emph{ESCO level 1}, 
which encompass general occupational categories. 
From \emph{ESCO level 5 onwards}, 
the taxonomy captures the most detailed occupations, 
with depths extending up to level 7. 
For instance, a \emph{Software Developer} (2512.4) is categorized at level 5, 
while a specialized role such as \emph{Import Export Specialist in Electronic and Telecommunications Equipment} (3331.2.1.11) is classified at level 7. 
This hierarchical structure can be likened to a tree, 
where \emph{3,039 unique occupations serve as leaf nodes}, 
representing the most specific classifications across different levels.

To map each job title to its corresponding ESCO code, 
we utilize a normalization function, 
leveraging the structured nature of ESCO data. 
Considering the critical importance of high-quality ESCO tagging for subsequent data analysis, 
we employed a state-of-the-art commercial ESCO tagger provided by Nobl.ai\footnote{\href{https://nobl.ai/}{https://nobl.ai/}}.
This commercial classifier utilizes embeddings of query job titles and job descriptions to retrieve the most likely ESCO codes.

\subsubsection{Validation}\label{sec_eval_data_normalization}

To evaluate the performance of the commercial classifier,
we hand-annotated 600 job experiences and their descriptions from the extracted data.
The annotation process for this dataset took approximately 30 hours and was conducted by the authors.
We also designed a simple evaluation pipeline, inspired by Li et al.~\cite{llm4jobs},
to compare the commercial classifier's performance with a baseline model.

\begin{figure}[htbp]
	\centering
	\resizebox{\columnwidth}{!}{
	   \includegraphics[width=0.83\linewidth]{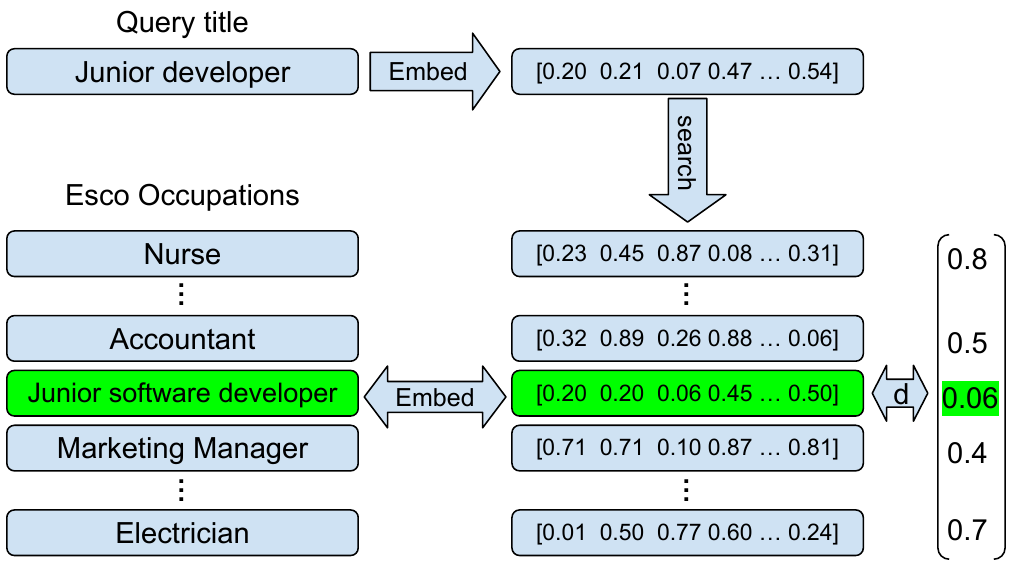}
	}
	\caption{High level diagram of how the function $f(\cdot)$ works. It takes a job title as input and returns the corresponding ESCO code. }
	\label{fig:mapping_diagram}
\end{figure}

The baseline normalization function is depicted in Fig.~\ref{fig:mapping_diagram}.
Initially, we used the Gemma 2 \cite{gemma2} model to convert all available ESCO codes into embeddings, 
which were then stored in a vector database.
Next, we embed the job titles of work experiences using the same sentence embedding model.
The cosine similarity between the embedding of each job title and the embedding of each ESCO code is then calculated, 
and the top ESCO codes with the highest similarity to the query are selected.

In Table~\ref{tab:embedding_comparison}, 
we compare the performance of the commercial classifier with the baseline model. 
The evaluation focuses on the models' ability to predict the correct ESCO code, ESCO level 4, and ESCO group.
The results show that the commercial classifier consistently outperforms the baseline across all ESCO levels. 
Despite the noisy and incomplete nature of the data, 
which limits the maximum achievable accuracy, 
the commercial classifier demonstrates strong performance on this task.
A qualitative analysis of the errors reveals that even when the classifier predicts the wrong ESCO code, 
the suggested codes are often semantically similar to the correct ones. 
This indicates that the produced data is sufficiently reliable for the downstream tasks.

\begin{table}[htbp]
\caption{Performance comparison between Nobl AI and the baseline across different ESCO levels.}
\begin{center}
\resizebox{\columnwidth}{!}{
\begin{tabular}{|c|c|c|c|}
\hline
\textbf{\textit{Model}} & \textbf{\textit{ESCO Group}} & \textbf{\textit{ESCO Level 4}} & \textbf{\textit{ESCO Code}} \\
\hline
Nobl AI & 84.7 & 77.8 & 72.6 \\
\hline
Baseline & 58.3 & 46.3 & 41.5 \\
\hline
\end{tabular}
}
\label{tab:embedding_comparison}
\end{center}
\end{table}

\subsubsection{Data Cleaning}\label{sec_cleaning}

As a sanity check in addition to the evaluation, 
we manually reviewed non-unique job titles with low matching scores, 
calculated via cosine similarity, 
to identify titles that were either too general or contained extraction errors. 
We created a shortlist of uninformative job titles, 
which was then expanded using semantic similarity models to include all potentially similar titles. 
As a final verification step, 
we employed ChatGPT-4o to further refine the list by assessing whether the job titles were too general or contained extraction errors.

General job titles are those from which employment status can be inferred but where the specific role is unclear, 
such as ``student job,'' ``employed,'' or ``holiday job.'' 
These titles were mapped to ``Unknown'' in the dataset. 
Titles with extraction errors, such as ``bachelor of \ldots'' or ``summer vacation,'' 
which do not correspond to actual occupation roles and accounted for over 1,200 rows, were excluded from the dataset.

Following these cleaning steps, we obtained 1,677,701 work experiences from 361,207 resumes. 
Of these, 206,261 work experiences were categorized as ``Unknown,'' which we retained in the dataset 
as they may indicate occupations for which date information is available. 
Not all work experiences in the dataset necessarily have a start or end date, 
since this information might not be provided in the resume or may have been removed during anonymization. 
Specifically, 1,226,842 work experiences were matched to an ESCO code and have at least one of these dates, 
while 1,046,925 experiences have both start and end dates. 
The resulting dataset, \emph{JobHop}, serves as the foundation for subsequent analyses in this study.

\newcommand{\idea}[1]{}
%===================================================
%===================================================
\section{Data Analysis}\label{sec_analysis}
%===================================================
%===================================================4

In Sec.~\ref{sec_processing}, we outlined the data processing steps used to create JobHop from raw resumes.
This section derives insights from the collected data by addressing various research questions, 
explored in Secs.\ref{sec:degree_effect}--\ref{sec:transitions}.

It is worth noting that the absence of data such as age, gender, and other demographic variables may limit certain types of analyses.
The analyses presented here are based on the final JobHop dataset, 
which contains ESCO codes, dates, and educational qualifications\footnote{A flag indicating whether the candidate obtained a tertiary degree} for each candidate.
These analyses demonstrate JobHop’s potential for studying labor market dynamics and deepen our understanding of employment patterns in Flanders, Belgium.

\idea{
\begin{itemize}
	\item How does holding a tertiary degree influence the types of jobs individuals pursue?
	\item How do career breaks impact subsequent job opportunities, and does this effect differ for individuals with a tertiary degree?
	\item How do job transitions vary across different ESCO groups?
	\item Which occupations tend to retain individuals for extended periods, and do durations differ for individuals with a tertiary degree? 
\end{itemize}
}

\emph{Remark:}
For clarity, 
Figures~\ref{fig:uni_effect}--\ref{fig:transition_matrix} and Table~\ref{tab:improved_duration} compare statistics across ESCO groups. 
Each ESCO group is annotated with a single descriptive word for brevity (e.g., “Army” for armed forces occupations, “Technicians” for Technicians and Associate Professionals). 
The full list of ESCO groups is provided alongside our published dataset.

\subsection{How does holding a tertiary degree influence the types of occupations individuals pursue? \label{sec:degree_effect}}

\begin{figure}[h]
    \centering
    \includegraphics[width=0.95\linewidth]{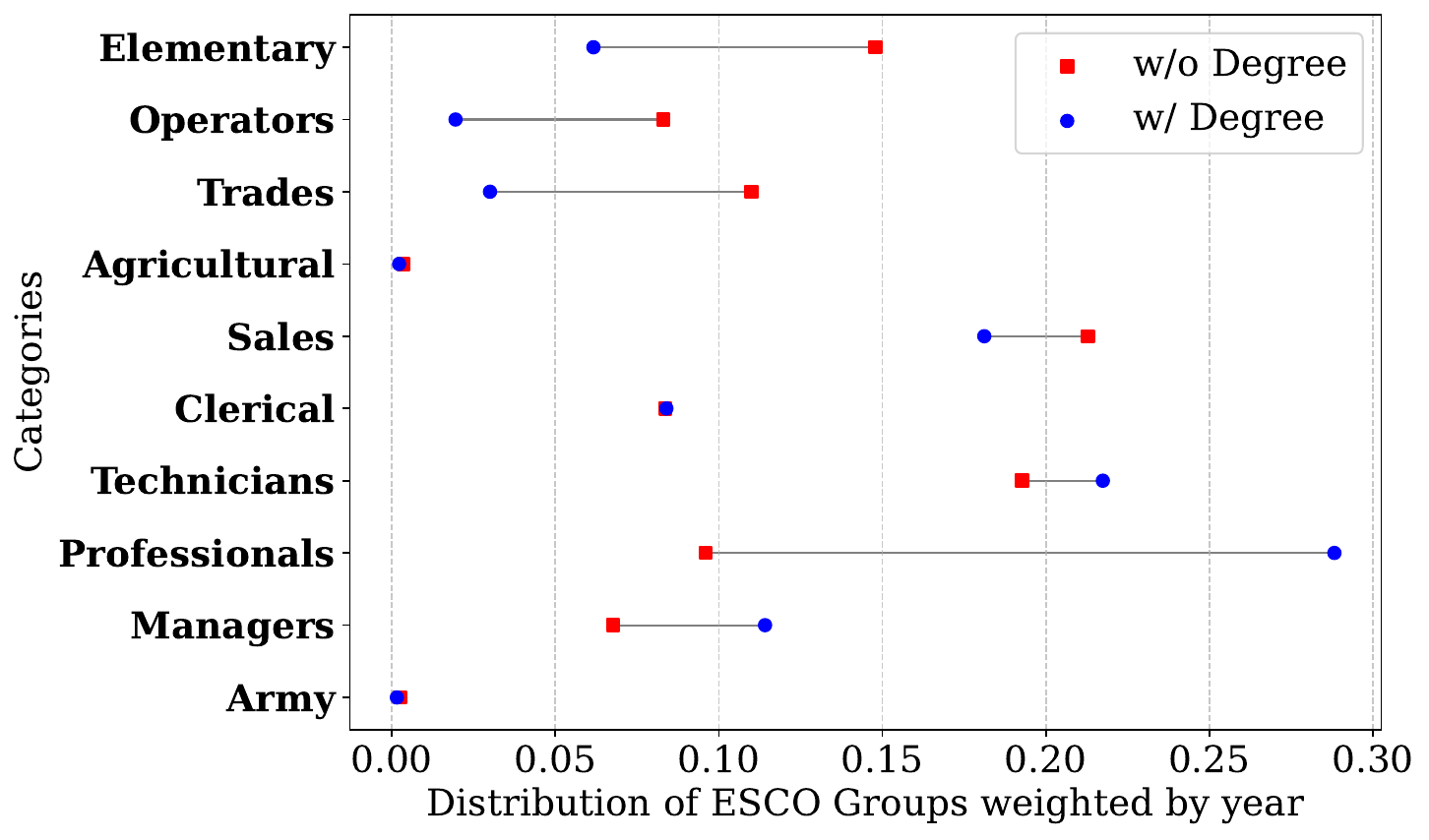}
    \caption{Normalized ratio of occupation category distribution between individuals with and without tertiary degrees, weighted by years in each occupation.}
    \label{fig:uni_effect}
\end{figure}

Figure~\ref{fig:uni_effect} presents a comparison of the normalized distribution of occupation categories for individuals with and without a tertiary degree. 
The distributions are weighted by occupation duration to account for varying employment lengths, 
with the length of the lines indicating the differences in occupation category distributions between the two groups.

Individuals with a tertiary degree are more likely to be employed in the “Professionals” and “Managers” categories than those without a degree,
with approximately 30 percent of degree holders employed in “Professionals” occupations that typically require specialized knowledge.
By contrast, individuals without a tertiary degree are more prevalent in “Elementary,” “Operators,” and “Trades” roles.
In occupations such as “Technicians,” “Clerical,” and “Sales,” 
there is no notable difference in the distribution between degree holders and non-degree holders,
suggesting that practical skills may outweigh formal education in these fields.

\subsection{How do career breaks impact subsequent occupation opportunities, and does this effect differ for individuals with a tertiary degree?\label{sec:break_effect}}

\begin{figure*}[htbp]
	\centering
	\includegraphics[width=\textwidth]{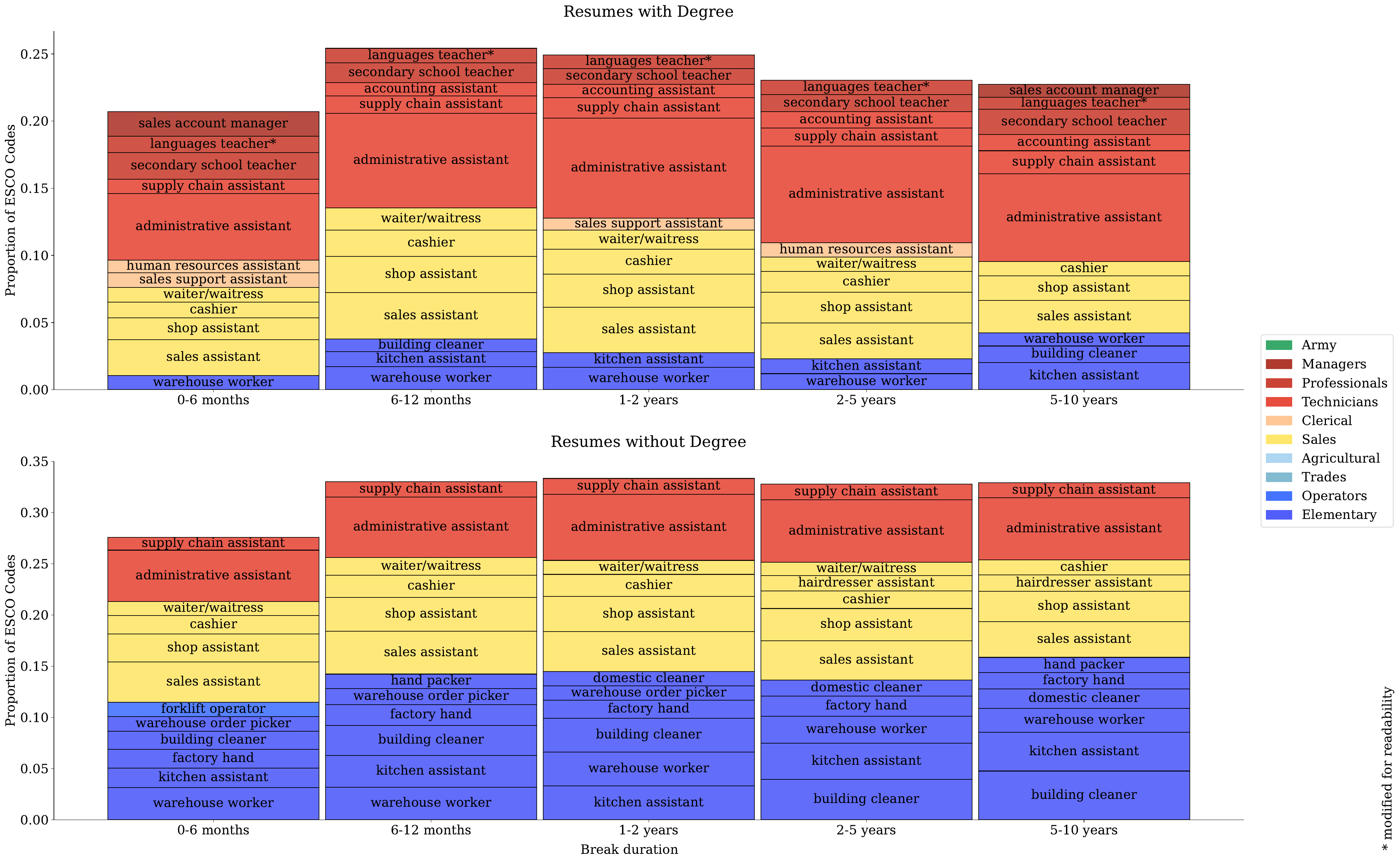}
	\caption{Fifteen most frequent post-career break employment outcomes by educational attainment. 
		The top panel represents resumes of individuals with a tertiary degree, 
		and the bottom panel represents resumes of individuals without a tertiary degree.}
	\label{fig:career_break_two}
\end{figure*}

\begin{figure*}[htbp]
	\centering
	\includegraphics[width=\textwidth]{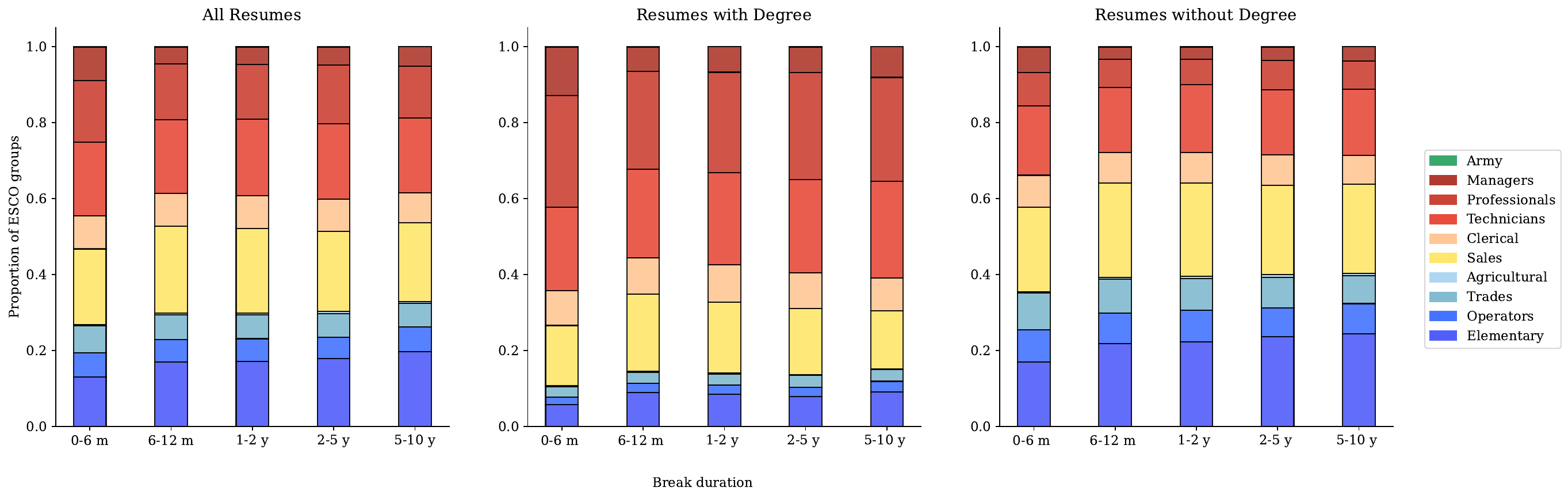}
	\caption{Percentage distribution of ESCO groups following a career break. The left panel shows all resumes, the middle panel focuses on individuals with a tertiary degree, and the right panel displays individuals without a tertiary degree.}
	\label{fig:break_groups_3}
\end{figure*}

Figures~\ref{fig:career_break_two} and \ref{fig:break_groups_3} illustrate the types of occupations individuals enter following a career break.
Career breaks are defined as periods during which no work experience is recorded.

In Fig.~\ref{fig:career_break_two}, each bar represents one of the 15 most frequent occupations individuals enter after a career break. 
The figure is divided into two panels by educational attainment,
with each panel displaying five bars for different break durations.
Bar heights indicate the share of individuals transitioning into each occupation.
The two bars on the right indicate occupations with high career-change frequency, 
while the remaining bars represent entry-level positions for individuals with breaks of at least one year.
Roles such as “Administrative Assistant,” “Sales Assistant,” and “Shop Assistant” are common post-break choices,
reflecting their prevalence in the dataset.
As break duration increases, lower-skill occupations, such as “Kitchen Assistant” and “Building Cleaner,” become more prevalent, 
reflecting the challenges of securing field-specific employment after a prolonged hiatus.

The top panel of Fig.~\ref{fig:career_break_two}, representing degree holders,
shows a higher prevalence of roles requiring advanced skills, such as “Accounting Assistant,” “School Teacher,” and “Sales Account Manager.” 
This suggests that tertiary education supports re-entry into field-related roles after a career break.
While the most frequent occupations are similar across both panels, positions requiring higher skill levels are more prevalent among degree holders,
underscoring the value of a tertiary degree in maintaining access to relevant roles despite career interruptions.

By contrast, the bottom panel, showing individuals without a tertiary degree, highlights a higher concentration in elementary occupations.
The 15 most frequent post-break occupations constitute a larger share for this group,
indicating fewer options after a break. 
The most frequent occupations include “Administrative Assistant,” “Sales Assistant,” “Building Cleaner,” “Warehouse Worker,” and “Kitchen Assistant.”

Figure~\ref{fig:break_groups_3} shows the percentage distribution across ESCO groups following a career break.
Unlike Fig.~\ref{fig:career_break_two}, this figure includes all occupations after a break, not just the 15 most frequent ones.
It is divided into three panels: the left shows all resumes,
the middle focuses on individuals with a tertiary degree,
and the right displays individuals without a tertiary degree.
Each panel shows five bars representing different break durations;
the height of each bar indicates the share of individuals entering each ESCO group post-break.

Across all panels in Figure~\ref{fig:break_groups_3}, as break duration increases, the share of lower-skill (“blue”) occupations grows, 
while the share of higher-skill (“red”) occupations declines, particularly for individuals without a tertiary degree.
For degree holders, the share of higher-skill occupations remains consistently larger, with slower declines as break duration lengthens.
These findings suggest that tertiary education helps individuals maintain access to higher-skill roles even after prolonged breaks.

\begin{figure*}
	\centering
	\includegraphics[width=\textwidth]{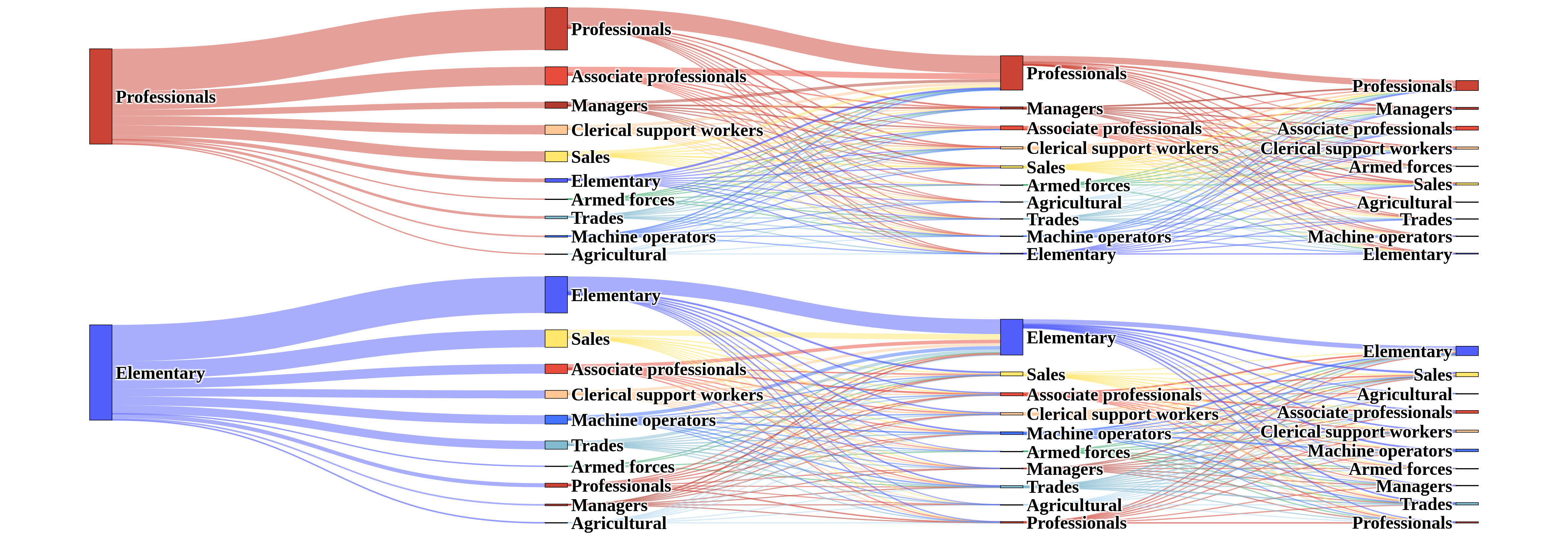}
	\caption{Occupation transitions for individuals starting in “Professional” and “Elementary” occupations.}
	\label{fig:sankey_transitions}
\end{figure*}

\begin{figure}[h!]
	\centering
	\includegraphics[width=0.83\linewidth]{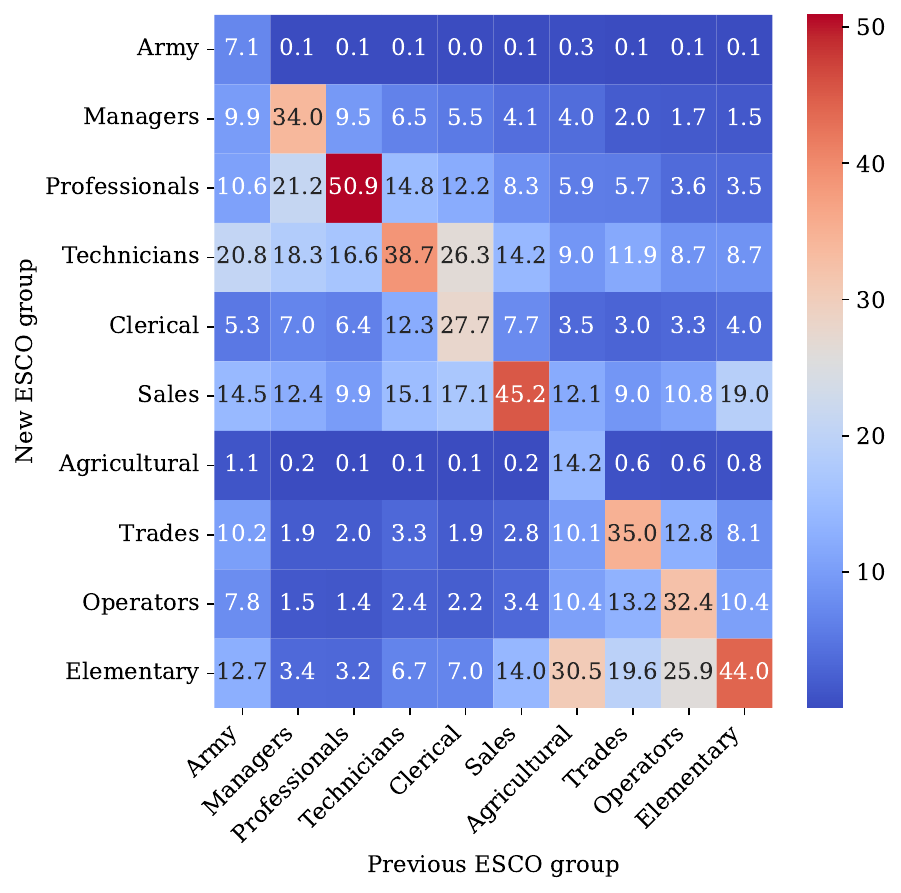}
	\caption{Transitions between different ESCO groups.}
	\label{fig:transition_matrix}
\end{figure}

\subsection{Which occupations tend to retain individuals for extended periods, and do durations differ for individuals with a tertiary degree?\label{sec:duration_analysis}}

Table~\ref{tab:improved_duration} reports the median and average occupation-spell durations (in years) for the top ten ESCO occupational groups, 
both overall and stratified by tertiary education status.
Most occupations exhibit a median tenure of around one year, reflecting the prevalence of shorter occupation spells.
Individuals without a tertiary degree tend to remain in the same occupation longer than degree holders; 
for example, non-degree managers have a median tenure of 2.50 years (average 4.16) versus 2.00 years (average 3.13) for degree holders.
A notable exception is agricultural occupations, where degree holders have a slightly higher median duration (1.25 vs.\ 1.00 years).
Among all groups, Elementary occupations have the shortest median and average durations,
indicating that these roles are often entry-level and prone to high turnover.
Finally, because the Army and Agricultural groups represent a relatively small share of the dataset (Fig.~\ref{fig:uni_effect}), \
their estimated durations may be subject to greater sampling variability and should be interpreted with caution.

\begin{table}[htbp]
\centering
\small
\begin{threeparttable}
\caption{Median and Average Job Duration (Years) by Occupation and Education Level}
\label{tab:improved_duration}
\setlength{\tabcolsep}{3pt}
\begin{tabular}{
    p{1.5cm}
    S[table-format=0.2] S[table-format=0.2]
    S[table-format=0.2] S[table-format=0.2]
    S[table-format=0.2] S[table-format=0.2]
}
\toprule
& \multicolumn{2}{c}{\textbf{All}} 
& \multicolumn{2}{c}{\textbf{Degree}} 
& \multicolumn{2}{c}{\textbf{No Degree}} \\
\cmidrule(lr){2-3} \cmidrule(lr){4-5} \cmidrule(lr){6-7}
\textbf{Occupation} & \textbf{Med.} & \textbf{Avg.} 
& \textbf{Med.} & \textbf{Avg.} 
& \textbf{Med.} & \textbf{Avg.} \\
\midrule
Army         & 2.00 & 3.51 & 2.00 & 2.96 & 2.00 & 3.73 \\
Managers     & 2.00 & 3.60 & 2.00 & 3.13 & 2.50 & 4.16 \\
Professionals& 1.00 & 2.60 & 1.00 & 2.26 & 2.00 & 3.40 \\
Technicians  & 1.08 & 2.87 & 1.00 & 2.30 & 1.67 & 3.35 \\
Clerical     & 1.00 & 2.73 & 1.00 & 2.14 & 1.42 & 3.17 \\
Sales        & 1.00 & 2.27 & 1.00 & 2.02 & 1.00 & 2.39 \\
Agricultural & 1.00 & 2.49 & 1.25 & 2.59 & 1.00 & 2.45 \\
Trades       & 1.41 & 3.26 & 1.00 & 2.43 & 1.59 & 3.41 \\
Operators    & 1.00 & 2.93 & 1.00 & 1.99 & 1.00 & 3.09 \\
Elementary   & 1.00 & 2.11 & 1.00 & 1.80 & 1.00 & 2.19 \\
\bottomrule
\end{tabular}
\begin{tablenotes}
\footnotesize
\item \textit{Note}: Medians and averages are in years, rounded to two decimal places.
\end{tablenotes}
\end{threeparttable}
\end{table}

To further explore which occupations retain workers the longest, 
Table~\ref{tab:duration_top10} lists the top ten occupations by median spell length for individuals with and without a tertiary degree,
limited to occupations with at least 500 observed spells for stability.\footnote{Ranking is by median; average durations illustrate skew.}

Table~\ref{tab:duration_top10} shows that certain non-degree occupations exhibit particularly long tenures 
(e.g., Case Administrator and Retail Entrepreneur, with medians exceeding four years), 
indicating substantial stability despite the absence of formal qualifications.
In contrast, the longest median tenure among degree holders reaches only three years 
(e.g., Animation Director and Commercial Director), 
suggesting that while professional roles are stable, 
turnover can be higher compared to some non-degree occupations,
possibly reflecting broader opportunities and higher mobility among degree holders.
Interestingly, managerial roles such as Trade Regional Manager and Business Manager appear in the top ten for both groups, 
yet non-degree holders in these roles often have longer tenures than their degree-holding counterparts.

Overall, while tertiary education offers various benefits, 
it does not consistently extend occupation tenure; 
some non-degree occupations retain workers longer than high-skill, degree-based roles. 
This pattern may reflect broader opportunities and higher mobility among degree holders.

\begin{table}[htbp]
\centering
\small
\begin{threeparttable}
\caption{Top Ten Jobs by Job Duration (Years), With and Without a Degree}
\label{tab:duration_top10}
\setlength{\tabcolsep}{6pt}
\renewcommand{\arraystretch}{1}
\begin{tabular}{p{0.2cm} |p{4.2cm} cc}
\toprule
\multicolumn{1}{c}{} & \textbf{Occupation} & \textbf{Med.} & \textbf{Avg.} \\
\midrule
\multirow{10}{*}{\rotatebox{90}{No Degree}} 
 & Case Administrator                   & 4.5 & 6.4 \\
 & Retail Entrepreneur                  & 4.2 & 6.8 \\
 & Production Supervisor                & 4.0 & 5.9 \\
 & Warehouse Manager                    & 3.0 & 4.6 \\
 & Bookkeeper                           & 3.0 & 5.2 \\
 & Purchaser                            & 3.0 & 5.1 \\
 & Business Manager                     & 3.0 & 5.2 \\
 & Assembly Line Operator		        & 3.0 & 6.8 \\
 & Middle Office Analyst                & 3.0 & 5.0 \\
 & Trade Regional Manager               & 3.0 & 4.5 \\
\midrule
\multirow{10}{*}{\rotatebox{90}{Degree}} 
 & Animation Director                   & 3.0 & 3.5 \\
 & Commercial Director                  & 3.0 & 4.4 \\
 & Trade Regional Manager               & 3.0 & 3.7 \\
 & Business Manager                     & 3.0 & 4.3 \\
 & Human Resources Manager              & 2.8 & 4.0 \\
 & ICT Project Manager                  & 2.5 & 3.6 \\
 & Medical Sales Representative         & 2.3 & 3.8 \\
 & Contact Centre Manager               & 2.2 & 3.4 \\
 & Sales Manager                        & 2.2 & 3.4 \\
 & Software Analyst                     & 2.2 & 4.4 \\
\bottomrule
\end{tabular}
\begin{tablenotes}
\footnotesize
\item \textit{Note}: Only occupations with $>500$ observed occupation spells are included.
\end{tablenotes}
\end{threeparttable}
\end{table}

\subsection{How do occupation transitions vary across different ESCO groups?\label{sec:transitions}}

Figure~\ref{fig:sankey_transitions} shows occupation transitions for individuals whose first occupation was classified as “Professional” or “Elementary,” 
the two most prominent groups in our dataset for both degree and non-degree holders.
The first bar shows the starting occupation, while the last bar shows the fourth occupation in their career.
For resumes with fewer than four occupations, only the available occupations are shown, which results in fewer occupations in the final bars.
As further supported by Fig.~\ref{fig:transition_matrix}, individuals tend to remain within the same ESCO group.
Interestingly, many individuals who change careers once tend to return to their original group if they change careers again.

Figure~\ref{fig:transition_matrix} presents the transition matrix across ESCO groups, 
considering two consecutive occupations as a transition only when the first occupation’s end date precedes the start date of the second.
Numbers are normalized by the total number of transitions from each group; each value shows the percentage of transitions to each group.
Few occupations from the “Army” and “Agricultural” groups appear in our dataset.
For other groups, the stronger red diagonal indicates that individuals are more likely to remain in the same group than to switch groups.
Notably, significant transitions occur between “Managers,” “Professionals,” and “Technicians.” 
These insights are particularly relevant for researchers and professionals working on job recommendation and career planning,
as the strong relationships between these categories could broaden the range of job recommendations and career pathways offered to job seekers.

\section{Conclusion and Future Work}\label{sec_conclusion}
%===================================================
%===================================================

In this paper, we introduced JobHop, 
a large-scale dataset of real-world occupational transitions derived from anonymized resumes provided by VDAB, 
the public employment service in Flanders, Belgium.
Using a two-step LLM-based pipeline, 
we extracted structured career information from unstructured resume data and mapped it to standardized ESCO occupation codes, 
resulting in a dataset of over 1.67 million ESCO occupations from more than 361,000 resumes.
By aligning career information with a standardized taxonomy, 
the dataset enables diverse labor market analyses, 
including career path prediction, workforce planning, and data-driven policy development.

To demonstrate the dataset’s potential, 
we performed an analysis of the Flemish labor market, 
exploring the relationship between education and job applications,
the impact of career breaks on job transitions, and sector-specific job stability.
Our results highlight the significant influence of tertiary education on career trajectories, 
with degree holders more likely to secure managerial and professional roles.
Additionally, while longer career breaks generally lead to transitions into lower-skill jobs, 
individuals with higher education face fewer difficulties to resume their careers in qualified jobs.
To support ongoing research, 
we have publicly released the aggregated dataset and plan to update it regularly to reflect advancements in LLM technology and integrate new data.

Future work will focus on enhancing data extraction and classification through advanced prompt engineering, 
fine-tuning, and the use of newer LLMs to improve the accuracy and granularity of career information.
Additionally, further analysis of the dataset can provide deeper insights into labor market dynamics, 
occupational mobility, and the long-term effects of career breaks and educational attainment.
Moreover, this dataset paves the way for career path prediction and data-driven labour market analysis.

We hope that JobHop will serve as a valuable resource for researchers, policymakers, and practitioners, 
enabling data-driven decision-making and contributing to the development of more effective labor market policies and career guidance tools.

%===================================================
%===================================================
%% Acknowledgements
%===================================================
%===================================================

% \section*{Acknowledgements}
% This research was funded by the ERC under the E.U.’s 7th Framework and H2020 Programs (ERC Grant Agreement no. 615517 and 963924),
% the Flemish Government (AI Research Program), the BOF (Bijzonder Onderzoeksfonds) of Ghent University (BOF20/DOC/144 and BOF20/IBF/117), and the FWO (Fonds Wetenschappelijk Onderzoek, Research Foundation Flanders; G0F9816N, 3G042220).

\bibliographystyle{./IEEEtran}
\bibliography{IEEEabrv, ./bibliography}

\begin{thebibliography}{10}
\bibitem{LLMDB}
``Lifetime labour market database,'' the LLMDB is derived from several administrative datasets linked together by a unique individual identifier, the national insurance number (NINo).
\end{thebibliography}

\end{document}